\documentclass{article}
\usepackage{spconf,amsmath,graphicx}
\usepackage{bbm}
\usepackage{hyperref}

\title{A Two-stage Framework for Compound Figure Separation}
\name{Weixin Jiang$^{1,2}$, Eric Schwenker$^{2,3}$, Trevor Spreadbury$^{2}$, Nicola Ferrier$^{4}$, Maria K. Y. Chan$^{2}$, Oliver Cossairt$^1$\thanks{This material is based upon work supported by Laboratory Directed Research and Development (LDRD) funding from Argonne National Laboratory, provided by the Director, Office of Science, of the U.S. Department of Energy under Contract No. DE-AC02-06CH11357. Use of the Center for Nanoscale Materials, an Office of Science user facility, was supported by the U.S. Department of Energy, Office of Science, Office of Basic Energy Sciences, under Contract No. DE-AC02-06CH11357. We gratefully acknowledge the computing resources provided and operated by the Joint Laboratory for System Evaluation (JLSE) at Argonne National Laboratory.}}
\address{$^1$Department of Computer Science, Northwestern University, USA\\
$^2$Center for Nanoscale Materials, Argonne National Laboratory, USA\\
$^3$Department of Materials Science and Engineering, Northwestern University, USA\\
$^4$Mathematics and Computer Science, Argonne National Laboratory, USA}
\begin{document}
%
\maketitle
\begin{abstract}
Scientific literature contains large volumes of complex, unstructured figures that are compound in nature (i.e. composed of multiple images, graphs, and drawings). Separation of these compound figures is critical for information retrieval from these figures. In this paper, we propose a new strategy for compound figure separation, which decomposes the compound figures into constituent subfigures while preserving the association between the subfigures and their respective caption components. We propose a two-stage framework to address the proposed compound figure separation problem.  In the first stage, the subfigure label detection module detects all subfigure labels. Then, in the subfigure detection module, the detected subfigure labels help to detect the subfigures by optimizing the feature selection process and providing the global layout information as extra features. Extensive experiments are conducted to validate the effectiveness and superiority of the proposed framework, which improves the  precision by 9\% compared to the benchmark.
\end{abstract}
\section{Introduction}
\label{sec:intro}

The ever-increasing volume of scientific literature, which remains as the authoritative data source for most scientific fields, necessitates efficient and effective information retrieval tools.  An obstacle towards automatic image data analysis and retrieval from scientific literature is that over 30\% of figures are compound (i.e. consist of more than one subfigure) in nature \cite{de2016overview}. Analyzing or retrieving information from these compound figures is challenging because applying an image feature detector to compound figures will result in a mixture (often an average) of features from different subfigures. However, the content of subfigures can vary dramatically or even be completely unrelated, which can reduce the discriminative power of image retrieval algorithms \cite{taschwer2018automatic}.

Therefore, decomposing compound figures into subfigures is a critical step towards information retrieval from these figures. Most previous work on compound figure separation decomposes the figures into the smallest possible subfigures \cite{yuan2014novel, lee2015detecting, taschwer2018automatic, tsutsui2017data, mukaddem2019imagedataextractor}, which misses an important opportunity to incorporate association between subfigures and caption information. It is important to note that captions provide essential description to help users understand the content of the figures in scientific articles. Frequently, authors include separate descriptions for each subcaption in the compound figure caption. Thus, a method for parsing compound figures into subfigures in such a way that is consistent with authorial intent has the promise to improve separation performance.

\begin{figure*}[htb]

\centering
\centerline{\includegraphics[width=2\columnwidth]{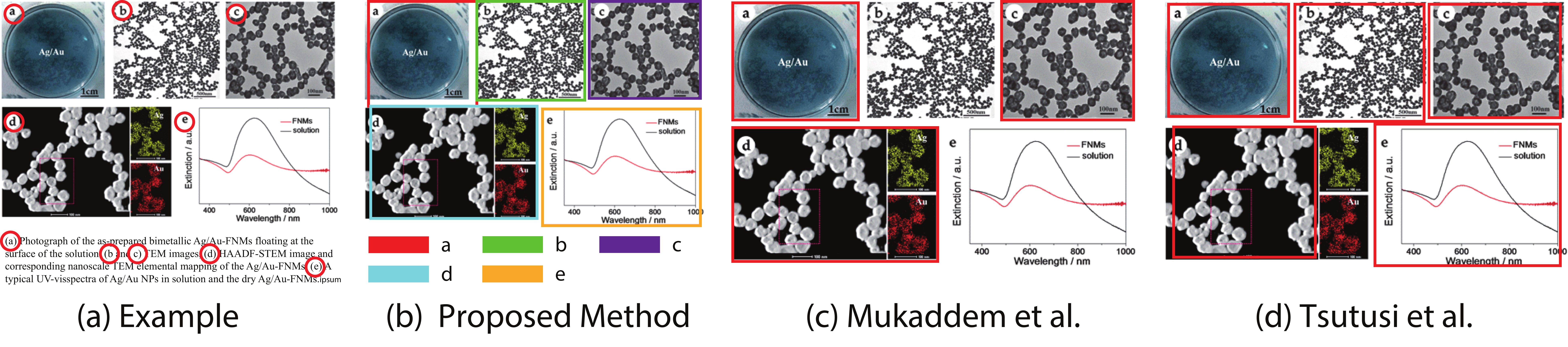}}
\caption{Comparison of different compound figure separation algorithms. (a). An example compound figure with its respective caption. Subfigure labels (the ones in red circles) appear in both the compound figure and the caption, bridging the connection between them. (b). The separation result by the proposed framework, which detects subfigures along with their respective subfigure labels. (c). The separation result by \cite{mukaddem2019imagedataextractor}. (d). The separation result by \cite{tsutsui2017data}. Image source: (Wu et al. Nanoscale, 2012)}
\label{fig:demo_comparison}
\end{figure*}

By convention, authors typically place indices (subfigure labels, i.e. "a", "b", "c", etc.) in front of the subfigures and their respective caption components\footnote{Our method also handles the case where sub-figure labels are placed above/below subfigures}, as shown in Fig. \ref{fig:demo_comparison}(a).  In this paper, we propose a new compound figure separation strategy that decomposes compound figures under the guidance of subfigure labels. In Fig. \ref{fig:demo_comparison}(b), we provide an example where the compound figure is decomposed into several subfigures while each subfigure is assigned a subfigure label. The decomposition is accomplished using a two-stage framework. In the first stage, subfigure labels are detected with the subfigure label detection module. Then in the second stage, the compound figure, along with layout information extracted from detected subfigure labels, are fed into the subfigure detection module for subfigure detection. 

\section{Related Work}
\label{sec:format}

\subsection{Compound figure separation}

The research community only began showing a concerted interest in the compound figure separation (CFS) problem when it was introduced as one of the ImageCLEF 2013 challenges in the biomedical domain   \cite{garcia2014overview}. In the past, hand-crafted features and rule-based methods were heavily utilized. Empty space between subfigures was used as an important feature to decompose compound figure \cite{yuan2014novel}, while \cite{lee2015detecting} et al. further merged split fragments by using a SVM-based classifier. Furthermore, Taschwer et al. \cite{taschwer2018automatic} segmented compound figures by taking both empty space and edges into consideration. Learning based methods, especially deep neural networks, were proposed later to solve the CFS problem. Tsutsui et al. \cite{tsutsui2017data} tackled the CFS problem with a conventional object detection network. Shi et al. \cite{shi2019layout} predicted the layout structure, along with the bounding boxes. However, only a limited number of predefined layout configurations are implemented in \cite{shi2019layout}, which prevents the algorithm from working effectively for general compound figures. None of the previous algorithms take subfigure labels into consideration, disregarding the connection between subfigures and their corresponding descriptions provided in compound figure captions. 

\subsection{Object detection}
Object detection algorithms aim at locating the position of objects in the target image and recognizing them. In past decades, the research community developed hand-crafted features to represent the target objects regardless of their scale, rotation, shift and illumination, such as Shift Invariant Feature Transform (SIFT) \cite{lowe2004distinctive} and Histogram of Oriented Gradients (HOG) \cite{dalal2005histograms}. More recently, deep neural networks have been introduced to address the object detection problem, leading to dramatically higher object detection performance on public benchmarks. The YOLO (You Look Only Once) object detection algorithm family \cite{redmon2016you, redmon2017yolo9000, redmon2018yolov3} addressed the problem with an end-to-end framework. In particular, YOLO \cite{redmon2016you} extracted a feature map from the target image and made predictions in a grid-by-grid manner. YOLOv3 \cite{redmon2018yolov3} further optimized the standard YOLO architecture by building the feature extractor with more convolutional layers and making predictions over different scaled feature maps. In the proposed framework, we build our subfigure label detection module and subfigure detection module on top of YOLOv3, in which the subfigure label placement serves as prior knowledge that is provided as input to preserve the association between different subfigures as well as to provide extra features.

\begin{figure}[htb]

\centering
\centerline{\includegraphics[width=0.65\columnwidth]{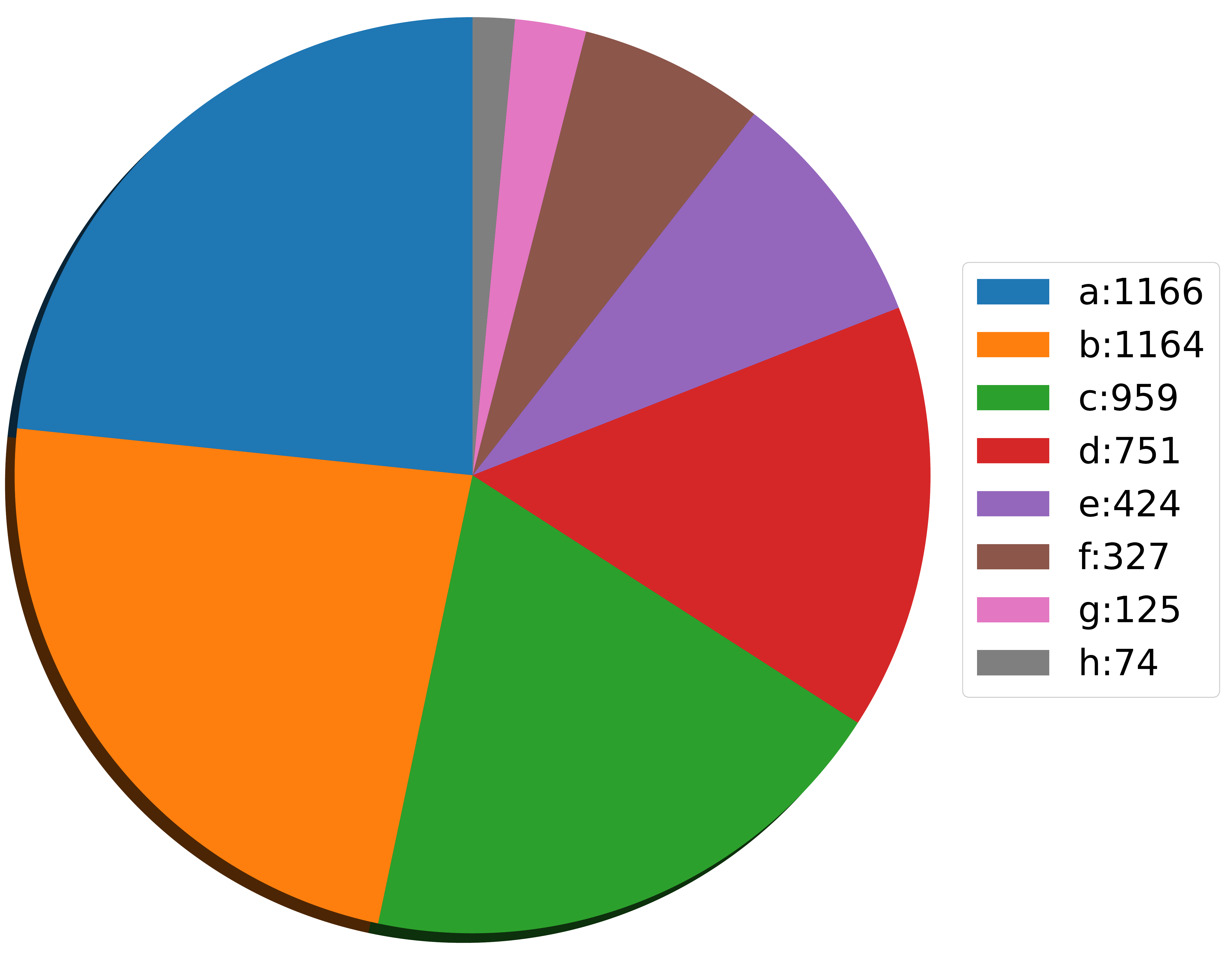}}
\caption{Distribution of different subfigure labels.}
\label{fig:dist_subfigure_labels}
\end{figure}

\begin{figure*}[htb]

\centering
\centerline{\includegraphics[width=1.8\columnwidth]{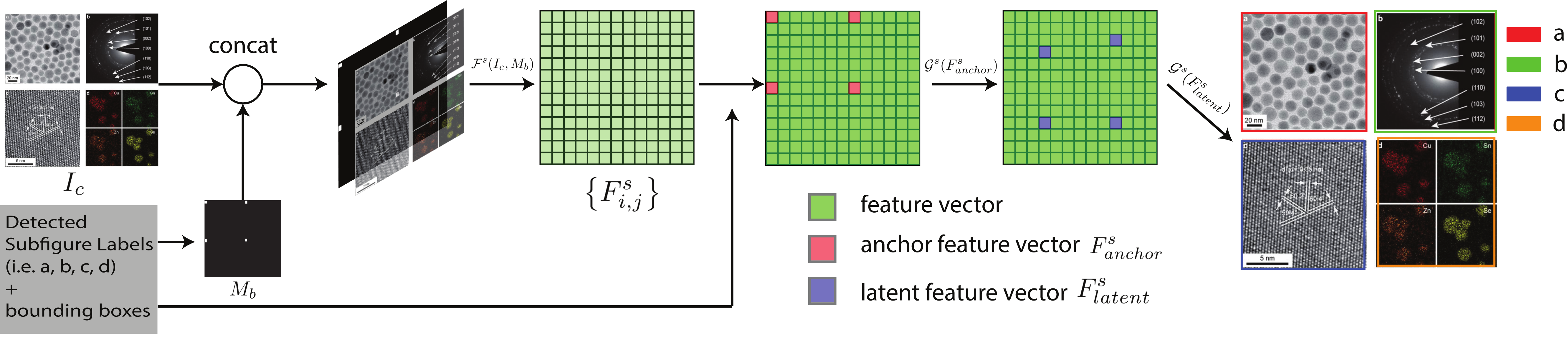}}
\caption{A step-by-step illustration of the subfigure detection module. Detected subfigure labels assist the detection of subfigures.}
\label{fig:demo_subfigure_label_detection_module}
\end{figure*}

\section{The approach}
Before we describe in detail the proposed framework, it is important to note that there are two existing challenges in addressing the compound figure separation problem. One is the imbalanced distribution of different subfigure labels (i.e. data imbalance problem). As shown in Fig. \ref{fig:dist_subfigure_labels}, we count the number of different subfigure labels in about 1200 compound figures. Obviously, some subfigure labels (e.g. "g", "h") occur far less frequently than  other subfigure labels (e.g. "a", "b"), which could easily lead to a subfigure label detector with a huge bias.  The other challenge is the huge in-class variation of subfigures. Target subfigures can sometimes be a combination of several images (e.g. the subfigure labeled by "d" in Fig. \ref{fig:demo_comparison}(a)). Previous methods, such as the CFS algorithm (e.g. \cite{tsutsui2017data} shown in Fig. \ref{fig:demo_comparison}(d)), will likely fail to make a correct detection because intrinsic features of the subfigure are not enough to lead to a confident prediction.

\subsection{Subfigure label detection module}
Unlike a conventional object detection framework, which trains the object localization module and recognition module simultaneously, we decouple these two modules to address the data imbalance problem. The subfigure label detection module is formulated as:
\begin{equation}
\begin{split}
 & \left\{F^l_{i,j}\right\} = \mathcal{F}^l(I_c) \\
 & B^l_{i,j}, O^l_{i,j} = \mathcal{G}^l(F^l_{i,j}) \\
 & C^l_{i,j} = \mathcal{H}^l(I_c, B^l_{i,j})\mathbbm{1}[O^l_{i,j} > \epsilon]
\end{split}
\label{eq: subfigure label detection module}
\end{equation}
where $\mathcal{F}^l$ is the feature extraction function, which encodes the input compound figure $I_c$ into a feature map, composing of a set of feature vectors $F^l_{i,j}$ ($i,j$ denote the coordinate of the feature map, $l$ denotes the subfigure label detection module). $\mathcal{G}^l$ is the bounding box prediction function, which predicts the bounding box $B^l_{i,j}$ and confidence score $O^l_{i,j}$ for each feature vector. $B^l_{i,j}$ is represented by the xy-coordinates of the center of the bounding box and the width and height of the bounding box (i.e. $B^l_{i,j} = \{x, y, w, h\}$). $\mathbbm{1}[.]$ is the indicator function. $\mathcal{H}^l$ is the classification function, which classifies the patches cropped from the compound figure into different subfigure label categories $C^l_{i,j}$.  During inference, low confidence bounding boxes are culled before being fed into the classification function.

Then the localization module (i.e. $\mathcal{F}^l, \mathcal{G}^l$) and classification module (i.e. $\mathcal{H}^l$) are optimized separately.

\begin{equation}
\begin{split}
  \mathcal{F}^l, \mathcal{G}^l &= \underset{\mathcal{F}^l, \mathcal{G}^l}{\arg\min}\sum_{i,j}\mathcal{L}_1(B^l_{i,j})+\lambda\mathcal{L}_2(O^l_{i,j}) \\
  \mathcal{H}^l &= \underset{\mathcal{H}^l}{\arg\min}\sum_{i,j}\mathcal{L}_3(C^l_{i,j}) \\
\end{split}
\label{eq: optim-subfigure label detection module}
\end{equation}
where $\mathcal{L}_1$ is the regression loss measuring the difference between the predicted bounding box and the ground truth. We use anchor-based loss\cite{redmon2017yolo9000,redmon2018yolov3} in this paper. $\mathcal{L}_2$ is the binary cross-entropy loss predicting the confidence scores and $\mathcal{L}_3$ is the cross-entropy loss which is widely used for classification problems. $\lambda$ is a hyperparameter.

\subsection{Subfigure detection module}
A key component in our framework is the subfigure detection module, which incorporates layout information from the detected subfigure labels to improve the performance of detecting subfigures within a compound figure. A step-by-step illustration of the  subfigure detection module is provided in Fig. \ref{fig:demo_subfigure_label_detection_module}. In particular, the layout information of the detected subfigure labels is encoded into a binary mask.

\begin{equation}
M_b(u,v) = \left\{
        \begin{array}{ll}
            1, & \quad (u, v) \in B^l \\
            0, & \quad \textrm{otherwise}
        \end{array}
    \right.
\label{eq: sum}
\end{equation}
where $B^l = \left\{B^l_{i,j} | C^l_{i,j} > 0\right\}$ is the set of bounding boxes of detected subfigure labels.

Then the binary mask is used as an extra feature, which is fed into the feature extraction function $\mathcal{F}^s$ along with the compound figure ($s$ denotes the subfigure detection module).
\begin{equation}
\left\{F^s_{i,j}\right\} = \mathcal{F}^s(I_c, M_b)
\label{eq: sum}
\end{equation}

Unlike conventional object detection frameworks, in which feature vectors with high confidence scores are selected, we use the location of the detected subfigure labels to select proper feature vectors (i.e. anchor feature vector $F^{s}_\textrm{anchor}$). By doing so, the association between subfigure labels and their respective subfigures are preserved. 
\begin{equation}
F^{s}_\textrm{anchor} = \left\{F^s_{x,y} | (x,y)\in B_{i,j}^l, C_{i,j}^l > 0  \right\}
\label{eq: sum}
\end{equation}

According to the anchor-based bounding box regression loss\cite{redmon2017yolo9000,redmon2018yolov3}, the feature vectors located at the center of the target object are more likely to make a more accurate estimation of bounding boxes. However, subfigure labels usually appear at the corner of their respective subfigures, thus the selected anchor feature vectors usually appear far from the center of the target subfigure. As shown in Fig. \ref{fig:demo_subfigure_label_detection_module}, we introduce the latent feature vectors (i.e. $F^{s}_\textrm{latent}$) for a more accurate bounding box prediction, which are computed from the anchor feature vectors but closer to the center of the target subfigures (see ablation study in Sec. \ref{sec: ablation study}). Then bounding boxes of subfigures are predicted from these latent feature vectors.
\begin{equation}
\begin{split}
B^{s}_a, O^{s}_a &= \mathcal{G}^s(F^{s}_\textrm{anchor}) \\
F^{s}_\textrm{latent} &= \left\{F^s_{x,y} | (x,y)\in B^s_a  \right\}\\
B^s, O^s &= \mathcal{G}^s(F^{s}_\textrm{latent})
\end{split}
\label{eq: sum}
\end{equation}
where $B^s, O^s$ denote the bounding box and confidence score of each detected subfigure. $B^{s}_a, O^{s}_a$ are auxiliary variables.

\begin{equation}
\mathcal{F}^s, \mathcal{G}^s = \underset{\mathcal{F}^s, \mathcal{G}^s}{\arg\min}\sum_{(x,y) \in B^s_a}\mathcal{L}_4(x,y)+\mathcal{L}_1(B^s)+\lambda\mathcal{L}_2(O^s)
\label{eq: optim-subfigure detection module}
\end{equation}
As shown in Eq. \ref{eq: optim-subfigure detection module}, the optimization process is based on three different loss functions, the regression loss (i.e. $\mathcal{L}_1$) of bounding boxes, the binary cross-entropy loss (i.e. $\mathcal{L}_2$) of confidence scores, and mean square error loss  (i.e. $\mathcal{L}_4$) measuring distance between the location of the latent feature vectors and the centers of subfigures.

\begin{table}
\begin{center}
\begin{tabular}{|c|c|c|c|}
\hline
Method & ${AP_{0.5}}$& ${AP_{0.75}}$& ${AP_{0.5:0.95}}$\\
\hline\hline
Mukaddem \cite{mukaddem2019imagedataextractor} & 24\% & 20\% & 19\% \\
Tsutusi \cite{tsutsui2017data}  & 84\% & 76\% & 63\%\\
Proposed &  90\% & 82\% & 72\%\\
\hline
\end{tabular}
\end{center}
\caption{Comparison of the performance of different compound figure separation methods.}
\label{tab: exp-performance comparison}
\end{table}



\begin{table}
\begin{center}
\begin{tabular}{|c|c|c|c|c|c|}
\hline
Methods & a & b & g & h & Avg.\\
\hline\hline
Baseline & 83\% & 80\% & 34\% & 40\% & 60\%\\
Proposed & 88\% & 89\% & 86\% & 81\% & 84\% \\


\hline
\end{tabular}
\end{center}
\caption{Comparison of the performance ($AP_{0.5}$) of training subfigure label detector with/without module decoupling.}
\label{tab: ablation study-decouple}
\end{table}


\section{Experiments}



Approximately 2000 compound figures were crawled from the Royal Chemistry Society (RCS), Springer Nature, and American Chemical Society (ACS)  journal families, and were uploaded to Amazon Mechanical Turk (MTurk), a platform for crowdsourced data labeling. Then the labeled figures were randomly divided into the training set ($\sim800$ figures) and the testing set ($\sim1200$ figures)\footnote{The code of this paper is released as part of the Github project: \url{https://github.com/MaterialEyes/exsclaim}}.

For the feature extraction function ($\mathcal{F}^l, \mathcal{F}^s$), we use Darknet-53\cite{redmon2018yolov3} to encode the input images into a stack of different scaled feature maps, followed by a feature pyramid layer\cite{lin2017feature}. We use 1x1 convolutional layer as the bounding box prediction function ($\mathcal{G}^l, \mathcal{G}^s$). We use the ResNet-152\cite{he2016deep} as the classification function (($\mathcal{H}^l$)) \footnote{A validation dataset is released on \url{https://petreldata.net/mdf/detail/exclaim_validation_v1.1/}}.

During the training process, we use the Adam\cite{kingma2014adam} optimizer while decaying the learning rate every 10000 iterations. It is worth noting that the classification function is trained with a mixture of real MTurk-labeled images and synthetic images. Each synthetic image is generated by cropping a random background patch from a random compound figure and pasting a random letter (e.g. "a", "b") onto it.

For the testing set containing 1164 images and 4982 subfigure labels, the detector detects 4777 true positive labels and 24 false positive labels, resulting in a precision equal to 99.5\% and a recall equal to 95.9\%.

As shown in Table. \ref{tab: exp-performance comparison}, the Average Precision (AP) is used to measure the performance of different compound figure separation.  Here the notation $AP_{\sigma}$ denotes the value of AP when setting the Intersect-over-Union (IoU) as $\sigma$. Mukaddem et al. \cite{mukaddem2019imagedataextractor} uses hand-crafted features for subfigure detection, an approach which works only for subfigure with sharp boundary and rich content (as shown in Fig. \ref{fig:demo_comparison}(c)). Tsutusi et al. \cite{tsutsui2017data} is a CNN-based method, which performs better than \cite{mukaddem2019imagedataextractor} but also fails to detect "compound subfigure" (as shown in Fig. \ref{fig:demo_comparison}(d)). As expected, with the help of subfigure labels, the proposed algorithm outperforms the other two methods, improving the mean Average Precision (mAP) metric by over 9\% compared to existing approaches.

\section{Ablation study}
\label{sec: ablation study}

We conduct two experiments here for ablation study.

One is to validate how the decoupling process addresses the data imbalance problem. As the baseline, we train the subfigure label detector directly with the training set. As shown in Table \ref{tab: ablation study-decouple}, we compare the $AP_{0.5}$ of the baseline and the proposed on the most frequent and least frequent subfigure labels. Without the decoupling process (Table 2), the performance drops significantly ($\sim40\%$) for the least frequent subfigure labels. As expected, the decoupling process produces a detector with less bias, improving average performance by 24\%. 

In another experiment, we validate how the latent feature vectors improve the detection precision. As a baseline, we predict the bounding boxes directly from the anchor feature vectors, which results in $AP_{0.5} = 90\%, AP_{0.75} = 65\%, AP_{0.5:0.95} = 60\%$. As expected, introducing the latent feature vectors significantly improves the detection precision (e.g. 17\%) with a large IoU threshold (e.g. $0.75$).


\section{Conclusion}

In this paper, we propose a new strategy to decompose compound figures into several subfigures while preserving the association between the subfigures and their respective subfigure labels. Meanwhile, we propose a two-stage framework to address the figure separation problem, which outperforms the state-of-the-art algorithms. 
\bibliographystyle{IEEEbib}
\bibliography{strings,refs}

\end{document}